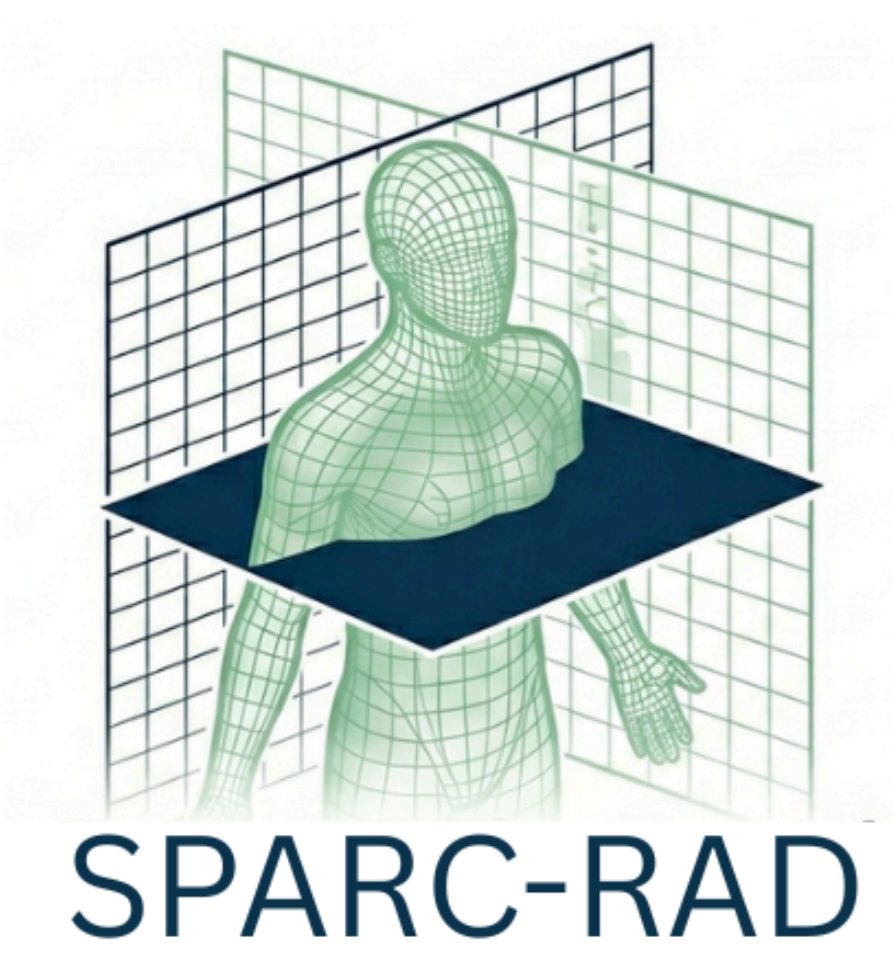

# SPARC-Rad: A Multimodal Benchmark Dataset and Evaluation Pipeline for Spatial and Anatomical Reasoning in Radiology Vision-Language Models

Satvik Tripathi[1*], Mustafa Ege Seker[2], Kristian Quevada[1.3], Ebubechukwu D Enwerem[1,4], Pratham Khandelwal[5], Emine Meltem[6], Bera Koca[7], Shahriar Faghani[1], Jacinta Arnold[1,8], Dania Daye[2], Tessa Cook[1]

1. Department of Radiology, Perelman School of Medicine, University of Pennsylvania, Philadelphia, Pennsylvania, USA
2. Department of Radiology, University of Wisconsin–Madison School of Medicine and Public Health, Madison, Wisconsin, USA
3. Department of Radiology, Cooper University Hospital, Cooper Medical School of Rowan University, Camden, New Jersey, USA
4. College of Computing and Informatics, Drexel University, Philadelphia, Pennsylvania, USA
5. Department of Computer Science and Engineering, University of Minnesota Twin Cities, Minneapolis, Minnesota, USA
6. Istanbul Training and Research Hospital, Department of Radiology, İstanbul, Türkiye
7. Department of Radiology, School of Medicine, Acıbadem Mehmet Ali Aydınlar University, Istanbul, Türkiye.
8. UC Davis Graduate School of Management, Davis, CA

*Corresponding-

satvik.tripathi@pennmedicine.upenn.edu
3400 Spruce st, Philadelphia PA, 19104

## Abstract

Vision-language models (VLMs) are increasingly being evaluated for medical imaging, but many available benchmarks emphasize disease classification, report generation, or broad visual question answering rather than the spatial and anatomical reasoning required for radiology. We developed the Spatial Perception and Anatomical Reasoning in Clinical Radiology (SPARC-Rad) Benchmark, a manually curated multimodal benchmark dataset and evaluation pipeline for assessing these capabilities in radiology VLMs. SPARC-Rad includes 300 image-question pairs derived from healthy control imaging studies in The Cancer Imaging Archive (TCIA), spanning CT, MRI, and radiography across the abdomen, chest, breast, neuro, and musculoskeletal categories. Radiology trainees manually designed and annotated questions to evaluate anatomical identification, localization, laterality, regional recognition, device identification, and inter-structure spatial relationships. The evaluation pipeline supports standardized prompting, structured output collection, response normalization, LLM-as-judge grading, human quality review, binary correctness scoring, and subgroup analysis by modality, anatomy, and reasoning type. SPARC-Rad provides a reusable framework for evaluating whether VLMs can provide reasoning for radiologic anatomy as a spatial system, supporting future model development, failure-mode analysis, and pre-deployment assessment.

## 1. Introduction

Vision-language models (VLMs) are increasingly being evaluated for medical imaging tasks, including image classification, visual question answering, report generation, and clinical decision support. Large public imaging resources such as MIMIC-CXR and CheXpert have supported important work in chest radiograph modeling, while structured resources such as RadGraph have enabled report-level relation extraction and evaluation [1], [2], [3], [4], [5]. However, radiology image understanding requires more than recognizing visual patterns or generating plausible text. Radiologists routinely interpret anatomy as a spatial system: they determine laterality, localize structures, reason across imaging planes, and describe relationships between organs, devices, and body regions.

Existing radiology visual question answering (VQA) resources have made major contributions to medical image-language research. VQA-RAD introduced clinically generated questions and answers about radiology images, while SLAKE expanded medical VQA through semantic labels, bilingual questions, and knowledge-enhanced annotations [6], [7]. These benchmarks are valuable, but they were not designed specifically to isolate spatial perception and anatomical reasoning as the primary construct being tested. As a result, a model may perform well on broad VQA or report-generation tasks while still making clinically important spatial errors, such as reversing laterality, mislocalizing a device, or confusing adjacent anatomical regions [8].

This limitation is increasingly important as general-purpose and medical VLMs are proposed for clinical, educational, and research use. General multimodal systems such as LLaVA showed the power of visual instruction tuning, while medical models such as Med-Flamingo, Med-PaLM Multimodal, and MedGemma reflect the rapid growth of domain-adapted medical VLMs [9], [10], [11]. Yet strong performance on broad multimodal tasks does not necessarily demonstrate reliable radiology-specific spatial understanding. Cross-sectional imaging, such as CT and MRI, requires reasoning about anatomy in three dimensions and sometimes in four dimensions. Radiography requires interpreting three-dimensional anatomy as projected onto a two-dimensional image. A benchmark designed to test these abilities must therefore include multiple modalities, diverse anatomical regions, and questions that directly probe spatial reasoning.

To address this gap, we developed Spatial Perception and Anatomical Reasoning in Clinical Radiology (SPARC-Rad), a manually curated multimodal benchmark for evaluating spatial perception and anatomical reasoning in radiology VLMs. In this manuscript, we describe the dataset design, construction pipeline, annotation strategy, evaluation workflow, statistical analysis framework, benchmark outputs, and recommended reporting practices. SPARC-Rad is a methodological framework for testing whether VLMs can reason about radiologic anatomy in a clinically meaningful and reproducible way, as supported by the data provided.

## 2. Dataset Design and Rationale

SPARC-Rad is designed as a focused benchmark for spatial perception and anatomical reasoning in radiology VLMs. The goal is not to test whether a model can diagnose disease or generate a complete radiology report; instead, the benchmark tests whether a model can identify imaged anatomic structures, understand their location, determine laterality, and interpret relationships between them. These skills are central to radiology practice but are often measured only indirectly in existing image-language benchmarks.

Each benchmark item uses an image-question pair that requires direct interpretation of the radiologic image. Questions may ask the model to identify a structure marked by an arrow, determine whether an object is left- or right-sided, localize a device, classify a body region, or describe the relationship between adjacent anatomical structures. This design evaluates whether the model can connect visual evidence to spatial and anatomical concepts, rather than relying only on memorized anatomy or general medical knowledge.

Healthy-control imaging was selected to reduce the confounding influence of abnormal pathology and to focus the benchmark on normal anatomy, spatial orientation, and modality-specific visual reasoning. This is important because failures in localization or laterality can occur even in the absence of disease. By beginning with healthy control studies, SPARC-Rad establishes a baseline assessment of anatomical and spatial understanding that can later be expanded to pathology-focused cases [1], [2].

As summarized in Tables 1 and 2, SPARC-Rad includes 300 image-question pairs across three imaging modalities and five major anatomical categories.

| Modality Categories | Count | Percentage |
|---|---|---|
| Radiography | 114 | 38.0% |
| CT | 98 | 32.7% |
| MRI | 88 | 29.3% |

**Table 1. SPARC-Rad dataset distribution by modality.**

| Anatomical Categories | Count | Percentage |
|---|---|---|
| Abdomen | 71 | 23.7% |
| Chest | 64 | 21.3% |
| Breast | 57 | 19.0% |
| Neuro | 54 | 18.0% |
| Musculoskeletal | 54 | 18.0% |

**Table 2. SPARC-Rad dataset distribution by anatomy.**

## 3. Benchmark Construction Pipeline

### 3.1 Image Retrieval and Source Selection

Source imaging was retrieved from healthy-control studies available through TCIA, an open-access research resource that de-identifies and hosts cancer imaging collections for public download and secondary research [1]. Healthy control studies were prioritized to assess normal anatomical understanding rather than to classify disease.

### 3.2 Multimodal Case Selection

Images were manually selected to represent CT, MRI, and radiography. The final dataset that we've provided to support our framework includes 98 CT image-question pairs, 88 MRI image-question pairs, and 114 XR image-question pairs. This distribution captures both cross-sectional and projection-based imaging.

### 3.3 Anatomical Category Selection

Cases were organized into five major anatomical categories: abdomen, chest, breast, neuro, and musculoskeletal imaging. The supporting dataset also includes more granular body regions and organ systems, including abdomen, pelvis, breast, chest, femur, hand, head and neck, pancreas, prostate, spine, and wrist. Both scout images and cross-sectional images were included for CT and MR. This diversity helps prevent overconcentration in a single region or modality and supports subgroup analysis by modality, anatomy, body region, and reasoning type.

### 3.4 Manual Question Design and Ground-Truth Annotation

For each selected image, radiology trainees (E.M. 5 years; B.K. 2 years radiology resident; M.E.S. 2 years radiology research fellow) manually designed questions intended to require image-grounded reasoning. Questions were constructed around anatomical identification, localization, laterality, regional identification, inter-structure spatial relationships, and counting or device recognition. Each image-question pair was assigned a concise reference answer. Because radiology terminology includes common abbreviations and synonyms, the benchmark supports equivalent-answer mapping, such as treating “IJ catheter” and “internal jugular venous catheter” as equivalent when laterality and localization remain correct.

Figure 1 shows a representative benchmark instance. The example illustrates how a radiograph can be paired with multiple spatially grounded questions and reference answers that test device identification, localization, and counting.

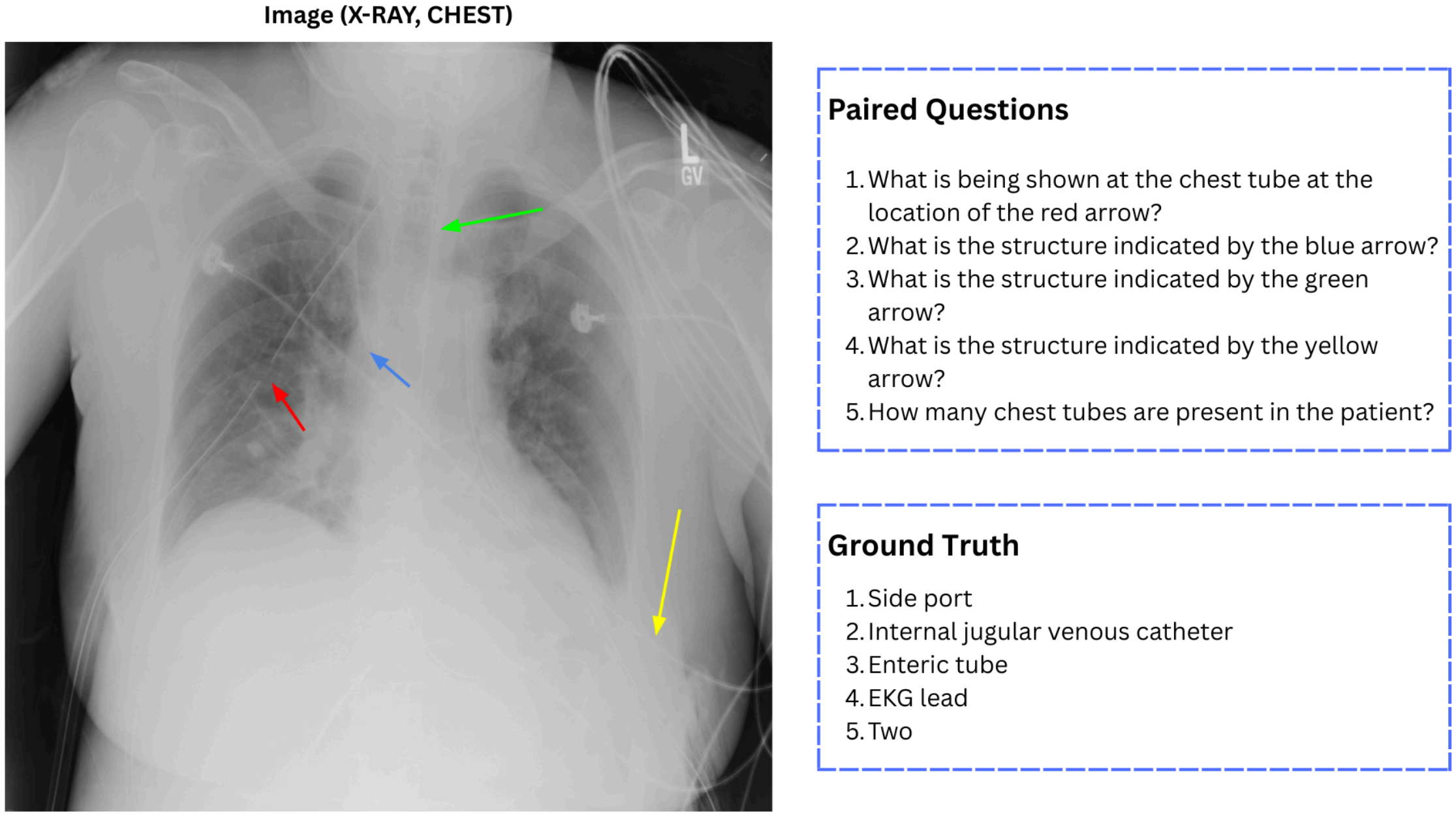

**Figure 1. Representative SPARC-Rad benchmark instance. A chest radiograph is paired with spatially grounded questions and corresponding reference answers to enable device identification and localization.**

## 3.5 Dataset Assembly and Quality Control

After image selection, question creation, and answer annotation, each benchmark instance was assembled as a structured image-question-answer record. Images were reviewed for visual clarity, modality labeling, and relevance to the intended anatomical category. Questions were reviewed to ensure that they were answerable from the image, spatially grounded, and not overly ambiguous. Ground-truth answers were checked for anatomical correctness and consistency. This quality-control process was essential because ambiguous questions or overly narrow reference answers can cause benchmark scores to reflect annotation noise rather than the model's true performance.

| Field | Description |
| --- | --- |

| Case_id | Unique identifier for each benchmark case |
|---|---|
| Image_id | Identifier for the selected radiologic image |
| Modality | CT, MRI, or radiography |
| Anatomical category | Abdomen, chest, breast, neuro, or musculoskeletal |
| Body_region | More specific anatomical region or organ system |
| Question_id | Unique identifier for each question |
| Question | Natural-language question provided to the model |
| Ground_truth_answer | Reference answer used for evaluation |
| Accepted_synonyms | Equivalent terms or abbreviations accepted during grading |
| Source_metadata | Source collection or dataset information when available |

**Table 3. Dataset schema for SPARC-Rad benchmark records.**

## 4. Evaluation Pipeline

SPARC-Rad is designed as both a benchmark dataset and a reproducible evaluation pipeline. Each image-question pair is provided to a candidate VLM using a standardized prompt. The model response is stored, normalized, graded against a reference answer, and then reviewed when needed. The resulting binary correctness labels support overall accuracy, subgroup analysis, model comparison, and error characterization.

Figure 2 summarizes the evaluation workflow. The pipeline begins with the SPARC-Rad image-question pair, collects a model output, compares it with the ground truth using answer normalization and LLM-as-judge grading, incorporates human review for quality control, and produces outputs for statistical analysis, subgroup analysis, and model comparison.

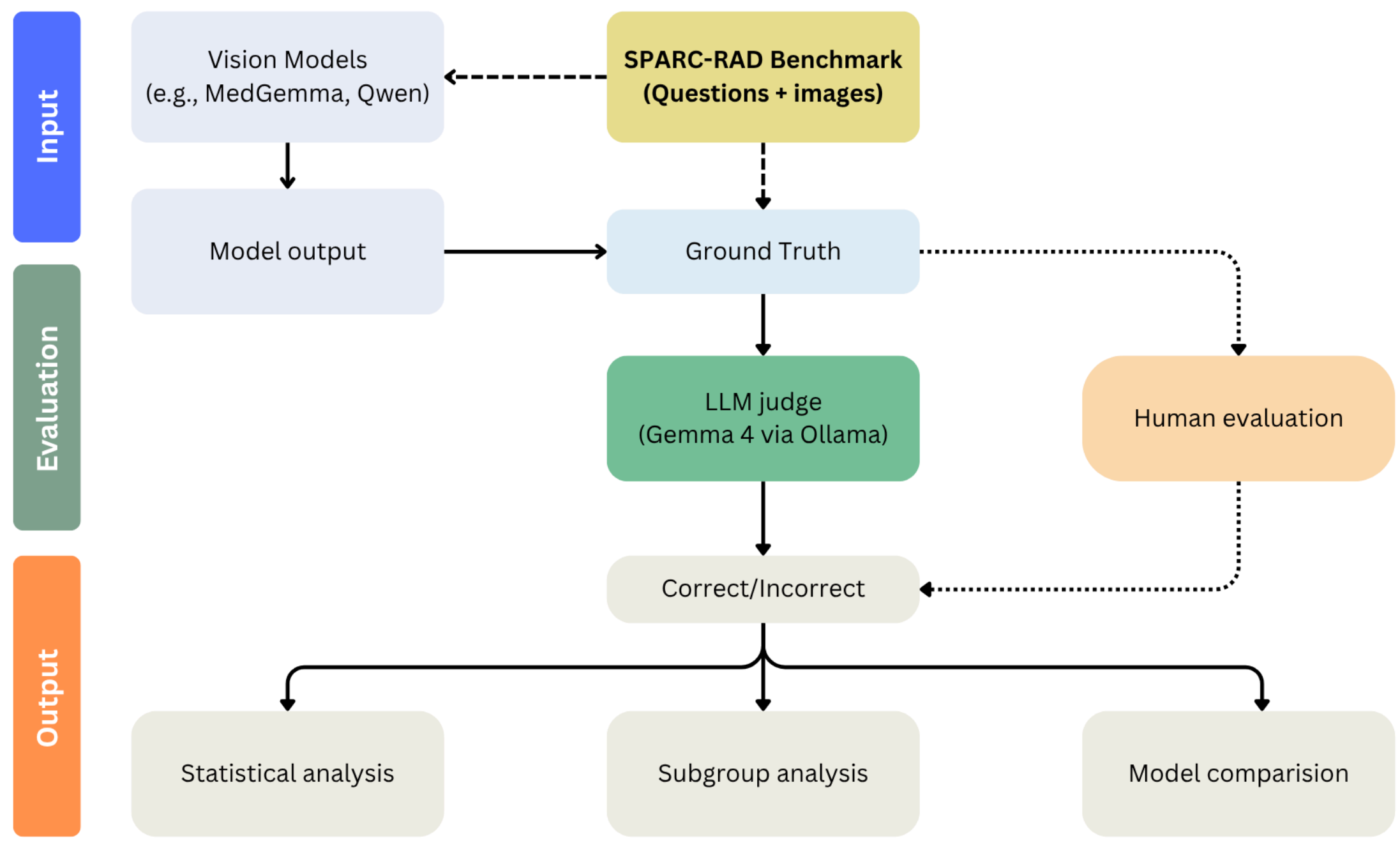

**Figure 2. SPARC-Rad evaluation pipeline. Candidate VLMs receive benchmark images and questions, produce free-text outputs, and are graded against reference answers using LLM-as-judge and human review before statistical analysis. Source: Authors' workflow diagram; contemporary medical VLM evaluation is informed by work on visual instruction tuning and medical VLMs [9], [11], [12].**

## 4.1 Model Input and Prompting

Each benchmark instance consists of a radiologic image and a spatially grounded question. The evaluation prompt was concise and instructed the model to answer only from the provided image. Our prompt was: "You are evaluating a radiology image. Answer the following question using only the provided image. Provide a concise answer. Question: [question text]." For reproducibility, each evaluation should record the exact prompt, model name, model version, inference date, decoding parameters, and raw model output.

## 4.2 Output Collection and Normalization

Model responses are stored before grading so that raw outputs remain auditable. Responses are then normalized to reduce false errors caused by wording differences. Normalization may include lowercasing, punctuation removal, abbreviation expansion, synonym mapping, and removal of extra explanatory language. Laterality terms should be standardized, but clinically meaningful distinctions must be preserved. For example, “left-sided” and “left side” may be mapped to “left,” but “left” and “right” should never be treated as interchangeable.

| Field | Description |
|---|---|
| Model name | Name of the evaluated model |
| Model version | Specific version or release |
| Case id | Unique case identifier |
| Question id | Unique question identifier |
| Prompt | Full prompt provided to the model |
| Raw response | Original model-generated answer |
| Parsed response | Cleaned or extracted final answer |
| Temperature | Decoding temperature, if available |
| Inference date | Date of model evaluation |
| Runtime environment | API, local deployment, or inference platform |

**Table 4. Recommended model-output schema for SPARC-Rad evaluations.**

### 4.3 Correctness Scoring, LLM-as-Judge Grading, and Human Review

The primary scoring outcome is binary correctness. A response is correct when it matches the reference answer or an accepted equivalent. A response is incorrect when it identifies the wrong structure, gives the wrong laterality, mislocalizes the object, provides a vague answer that does not resolve the question, or includes contradictory information. Because radiology answers often include acceptable synonyms, exact string matching alone is insufficient. SPARC-Rad therefore supports an LLM-as-judge approach in which a separate judging model receives the original question, reference answer, model response, and accepted synonyms, then determines whether the response is anatomically and spatially equivalent to the ground truth.

Human review is incorporated for ambiguous answers, to assess model uncertainty, to resolve disagreements between exact matching and LLM-based grading, and in cases where a response is partially correct but spatially incomplete. This review is especially important for laterality, device localization, and inter-structure relationships, where a response may contain related terminology yet still be clinically incorrect. The final output is a structured results file containing one row per model-response-question pair with metadata, raw response, normalized response, ground truth, and correctness label.

## 5. Statistical Analysis Framework

SPARC-Rad supports both overall model evaluation and subgroup-level analysis. The primary metric is overall accuracy, defined as the proportion of the 300 image-question pairs answered correctly. Each model should be reported with a 95% confidence interval to convey uncertainty and to avoid overinterpreting small differences in leaderboard rank. Because the benchmark includes structured metadata, accuracy should also be reported by modality, anatomical category, body region, and reasoning type.

When multiple models are evaluated on the same question set, pairwise comparisons should be interpreted carefully. Chi-square testing can compare distributions of correct and incorrect responses between models. However, McNemar testing is especially appropriate for matched question sets because it focuses on discordant cases where one model is correct, and the other is incorrect [13]. This approach helps determine whether observed differences reflect meaningful paired performance differences rather than only small changes in overall accuracy.

| Reasoning Type | Description |
|---|---|
| Anatomical identification | Identifying a structure, organ, device, or region |
| Localization | Determining where a structure or object is located |
| Laterality | Distinguishing left-sided from right-sided anatomy |
| Regional classification | Identifying the broader body region or imaging field |
| Inter-structure relationship | Describing the spatial relationship between structures |
| Counting/device recognition | Counting visible devices or identifying multiple objects |

**Table 5. Reasoning-type categories for subgroup analysis.**

In addition to accuracy, SPARC-Rad can assess category-level dispersion. Models with similar overall performance may differ in consistency across modality or anatomy. For example, a model with high abdomen accuracy but low chest radiograph accuracy may be less reliable than a model with the same average accuracy but more balanced subgroup performance. Recommended dispersion measures include standard deviation across categories, the range between best and worst category, minimum subgroup accuracy, and macro-average accuracy.

| Error Type | Example |
|---|---|
| Laterality error | Right instead of left |
| Wrong structure | Catheter mistaken for EKG lead |
| Wrong region | Abdomen instead of pelvis |
| Overly vague response | "Tube" instead of "enteric tube" |
| Relational error | Incorrect medial/lateral or superior/inferior relationship |
| Modality misunderstanding | Treating a radiograph like a cross-sectional image |
| Hallucinated finding | Describing a structure not visible in the image |

**Table 6. Common error types for qualitative failure-mode analysis.**

## 6. Results

The final SPARC-Rad benchmark consists of 300 radiologic image-question pairs. Each instance includes a radiologic image, a manually designed spatially grounded question, a reference answer, and structured metadata describing the modality, anatomical category, body region, and reasoning task. This structure allows SPARC-Rad to be used as a standalone benchmark and as a reusable evaluation pipeline for comparing model performance across clinical imaging domains.

As shown in Figure 3, radiography accounts for the largest share of the dataset, with 114 image-question pairs. CT contributes 98 pairs, and MRI contributes 88 pairs. This modality balance allows the benchmark to assess whether VLMs generalize across cross-sectional and projection-based representations. The anatomical distribution includes abdomen, chest, breast, neuro, and musculoskeletal imaging, enabling analysis across domains with different spatial demands.

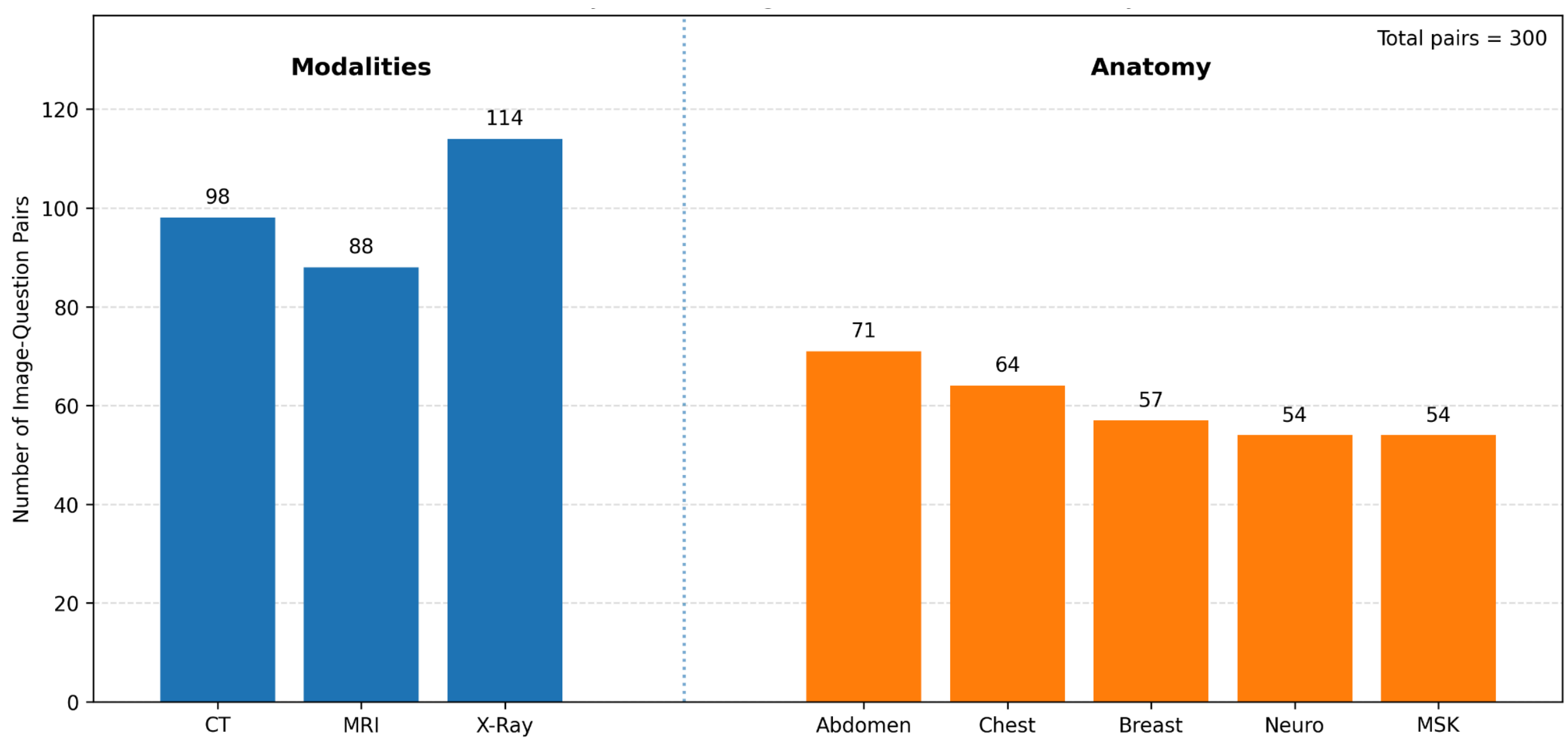

**Figure 3. Distribution of 300 manually curated image-question pairs from healthy-control TCIA studies.**

SPARC-Rad also supports several benchmark outputs: overall model accuracy, modality-stratified accuracy, anatomical category accuracy, reasoning-type accuracy, pairwise model comparison, and qualitative error analysis. These outputs are intended to move beyond a single leaderboard score and provide a more clinically meaningful profile of model behavior.

| Reporting Element | Purpose |
|---|---|
| Overall accuracy with 95% CI | Primary performance summary |
| Modality-stratified accuracy | CT, MRI, and radiography performance |
| Anatomical category accuracy | Abdomen, chest, breast, neuro, and musculoskeletal performance |
| Reasoning-type accuracy | Specific spatial reasoning skill assessment |
| Pairwise model comparison | Statistical testing between models |
| Error analysis | Identification of clinically meaningful failure modes |
| Model and inference details | Reproducibility and version tracking |

**Table 7. Minimum recommended reporting elements for SPARC-Rad evaluations.**

## 7. Discussion

SPARC-Rad was developed to address a specific gap in radiology vision-language model evaluation: the limited assessment of spatial perception and anatomical reasoning as distinct model capabilities. Existing radiology benchmarks frequently emphasize disease recognition, report generation, or broad visual question answering. Although these tasks are clinically relevant, strong performance on them does not necessarily demonstrate that a model can reliably identify anatomical structures, determine laterality, localize devices, or interpret spatial relationships. SPARC-Rad therefore isolates these foundational skills through image-grounded questions designed around anatomical identification, localization, laterality, regional recognition, device identification, counting, and inter-structure relationships.

Several methodological choices were made to support this focused objective. First, the initial benchmark was constructed using healthy-control imaging. This decision reduces the influence of pathology, postoperative change, and anatomical distortion, allowing performance to more directly reflect understanding of normal radiologic anatomy and spatial orientation. Establishing this baseline is important because failures in laterality or localization may occur even when no abnormality is present. However, this design also limits the benchmark's immediate ability to assess spatial reasoning in clinically complex cases. Model performance on SPARC-Rad should therefore be interpreted as a measure of foundational anatomical reasoning rather than comprehensive diagnostic competence.

Second, SPARC-Rad includes CT, MRI, and radiography because these modalities impose different spatial reasoning demands. CT and MRI require interpretation of sectional anatomy, imaging planes, organ position, and relationships across adjacent structures. Radiography instead compresses three-dimensional anatomy into a two-dimensional projection, creating challenges related to overlap, orientation, and apparent spatial relationships. Including both cross-sectional and projection-based imaging allows the benchmark to determine whether a model's reasoning generalizes across visual representations. At the same time, performance differences between modalities may reflect several interacting factors, including image complexity, model pretraining exposure, resolution, and modality-specific representation. Modality-stratified results should therefore be reported rather than relying only on a single overall accuracy value.

Third, the benchmark was designed to span abdomen, chest, breast, neuro, and musculoskeletal imaging. This anatomical breadth reduces dependence on any single clinical domain and enables identification of region-specific weaknesses that may be hidden within aggregate performance. A model may achieve a competitive overall score while performing poorly on laterality in musculoskeletal imaging, device localization on chest radiographs, or regional recognition in cross-sectional imaging. For this reason, SPARC-Rad is intended to produce a multidimensional performance profile rather than function solely as a leaderboard. Accuracy by modality, anatomical category, body region, and reasoning type is likely to be more informative than model rank alone.

The use of manually created questions was another deliberate design choice. Radiology trainees constructed image-grounded questions to target clinically meaningful spatial tasks that may not be captured through automatically generated question-answer pairs. Manual annotation allows questions to be tailored to the visual content and intended reasoning construct. However, it may also introduce variability in question difficulty, wording, specificity, and annotation style. Although questions and reference answers underwent quality review, additional board-certified radiologist review, multi-reader adjudication, and formal assessment of inter-rater agreement would further strengthen ground-truth reliability. Future benchmark versions should also consider calibrated difficulty levels and more explicit operational definitions for each reasoning category.

Free-text response evaluation introduces an additional methodological challenge. Exact string matching would incorrectly penalize many clinically equivalent answers because radiology terminology includes abbreviations, synonyms, and varying levels of specificity. SPARC-Rad therefore combines answer normalization, accepted-synonym mapping, LLM-as-judge grading, and human review. This approach increases flexibility and scalability, but it may also introduce grader dependence. Judging models may vary in how they interpret partial answers, terminology, or clinically meaningful distinctions. Human review remains particularly important for laterality, device localization, and relational questions, where a response may contain relevant language while still being spatially incorrect. Future evaluations should report the judging model, prompt, version, disagreement rate, adjudication process, and agreement between automated and human grading.

The primary outcome is binary correctness because it provides a clear and reproducible measure of whether the response resolves the spatial question. Nevertheless, binary scoring does not fully represent partial understanding. A model may correctly identify a structure while assigning the wrong laterality, or provide the correct anatomical region without sufficient specificity. Such responses are appropriately classified as incorrect under the current framework because the omitted spatial information may be clinically consequential. However, future versions could supplement binary scoring with structured error labels or hierarchical scoring that distinguishes identification, localization, laterality, and relational accuracy.

SPARC-Rad is also limited by its current size and modality coverage. The 300 image-question pairs provide an initial test set but may not fully represent the diversity of acquisition protocols, imaging planes, institutions, scanners, and patient populations encountered in clinical practice. Ultrasound, fluoroscopy, nuclear medicine, pediatric imaging, and longitudinal imaging are not included. The benchmark also does not currently assess pathological distortion, postoperative anatomy, artifacts, or uncommon anatomical variants. Expansion into these areas will be necessary to determine whether model performance remains stable under more challenging and clinically realistic conditions.

Potential data contamination must also be considered when evaluating general-purpose and medical VLMs. Although the questions were manually developed, the source imaging was obtained from public TCIA collections. Models may have encountered related images or collection metadata during pretraining, even if they have not encountered the specific image-question pairs. Performance should therefore not be interpreted as evidence of complete out-of-distribution generalization. Controlled-access test sets, hidden evaluation servers, or the inclusion of institutionally sourced external validation cases could reduce this concern in future studies.

Finally, SPARC-Rad should not be considered a substitute for clinical validation. A model that performs well on normal-anatomy spatial questions may still fail in diagnostic interpretation, clinical communication, workflow integration, or safety-critical settings.

Conversely, poor performance on particular SPARC-Rad subgroups may help identify foundational weaknesses before a model is considered for educational or clinical use. The benchmark is therefore best positioned as one component of a broader evaluation strategy that includes pathological cases, prospective testing, external validation, bias assessment, workflow studies, and post-deployment monitoring.

Future work should expand the number and diversity of cases, incorporate pathology-focused and postoperative questions, include additional modalities, and increase expert adjudication. Standardized evaluation scripts and hidden test sets would improve reproducibility and reduce contamination risk. Further analyses should examine whether errors cluster by modality, anatomical region, imaging plane, prompt wording, question difficulty, or reasoning type. These investigations may help distinguish general visual limitations from clinically specific spatial reasoning failures and guide targeted model development.

## 8. Conclusion

SPARC-Rad is a manually curated multimodal benchmark and evaluation pipeline for assessing spatial perception and anatomical reasoning in radiology VLMs. Its 300 image-question pairs test foundational capabilities across CT, MRI, and radiography, including anatomical identification, localization, laterality, regional recognition, device identification, and spatial relationships.

Rather than relying only on overall accuracy, SPARC-Rad supports analysis by modality, anatomy, and reasoning type, enabling clinically meaningful characterization of model strengths and failure modes. Although the current benchmark focuses on healthy-control imaging and requires broader expert validation and external expansion, it provides a structured foundation for evaluating whether VLMs can reason reliably about radiologic anatomy before clinical or educational deployment.